\documentclass{article}

\usepackage{microtype}
\usepackage{graphicx}
\usepackage{subfigure}
\usepackage{booktabs} 

\usepackage{hyperref}

\usepackage[accepted]{icml2024}

\usepackage{amsmath}
\usepackage{amssymb}
\usepackage{mathtools}
\usepackage{amsthm}
\usepackage{multirow}
\usepackage{adjustbox}

\usepackage[capitalize,noabbrev]{cleveref}

\theoremstyle{plain}
\newtheorem{theorem}{Theorem}[section]
\newtheorem{proposition}[theorem]{Proposition}

\theoremstyle{definition}

\newtheorem{assumption}[theorem]{Assumption}
\theoremstyle{remark}

\usepackage[textsize=tiny]{todonotes}

\icmltitlerunning{Fair Resource Allocation in Multi-Task Learning}

\begin{document}

\twocolumn[
\icmltitle{Fair Resource Allocation in Multi-Task Learning}



\icmlsetsymbol{equal}{*}

\begin{icmlauthorlist}
\icmlauthor{Hao Ban}{yyy}
\icmlauthor{Kaiyi Ji}{yyy}
\end{icmlauthorlist}

\icmlaffiliation{yyy}{Department of Computer Science and Engineering, University at Buffalo, New York, United States}

\icmlcorrespondingauthor{Kaiyi Ji}{kaiyiji@buffalo.edu}

\icmlkeywords{Multi-Task Learning, Fair Resource Allocation, Utility Maximization}

\vskip 0.3in
]



\printAffiliationsAndNotice{} 

\begin{abstract}
By jointly learning multiple tasks, multi-task learning (MTL) can leverage the shared knowledge across tasks, resulting in improved data efficiency and generalization performance. However, a major challenge in MTL lies in the presence of conflicting gradients, which can hinder the fair optimization of some tasks and subsequently impede MTL's ability to achieve better overall performance. Inspired by fair resource allocation in communication networks, we formulate the optimization of MTL as a utility maximization problem, where the loss decreases across tasks are maximized under different fairness measurements. To address the problem, we propose FairGrad, a novel optimization objective. FairGrad not only enables flexible emphasis on certain tasks but also achieves a theoretical convergence guarantee. Extensive experiments demonstrate that our method can achieve state-of-the-art performance among gradient manipulation methods on a suite of multi-task benchmarks in supervised learning and reinforcement learning. Furthermore, we incorporate the idea of $\alpha$-fairness into the loss functions of various MTL methods. Extensive empirical studies demonstrate that their performance can be significantly enhanced. Code is available at \url{https://github.com/OptMN-Lab/fairgrad}.
\end{abstract}

\section{Introduction}
\label{introduction}

By aggregating labeled data for various tasks, multi-task learning (MTL) can not only capture the latent relationship across tasks but also reduce the computational overhead compared to training individual models for each task \cite{caruana1997multitask,evgeniou2004regularized,thung2018brief}. As a result, MTL has been successfully applied in various fields like natural language processing \cite{liu2016recurrent,zhang2023survey,radford2019language}, computer vision \cite{zhang2014facial,dai2016instance,vandenhende2021multi}, autonomous driving \cite{chen2018multi,ishihara2021multi,yu2020bdd100k}, and recommendation systems \cite{bansal2016ask,li2020improving,wang2020m2grl}. Research has shown that MTL is capable of learning robust representations, which in turn helps avoid overfitting certain individual tasks \cite{lounici2009taking,zhang2021survey, ruder2017overview,liu2016algorithm}, and hence often achieves better generalization than the single-task counterparts.

MTL often solves the average loss across tasks in many real-world scenarios. However, it has been shown that there may exist conflicting gradients \cite{yu2020gradient,liu2021conflict,wang2020gradient,sener2018multi} among tasks that exhibit different directions and magnitudes. If directly optimizing the average loss, the final update direction will often be dominated by the largest gradient, which can degrade the overall performance of MTL. To alleviate this negative impact, a series of gradient manipulation methods have been proposed to find a compromised direction~\cite{desideri2012multiple, chen2018gradnorm, yu2020gradient, liu2021conflict, navon2022multi, xiao2023direction, liu2023famo}. In this paper, we view these methods from a novel fairness perspective. For example, MGDA \cite{desideri2012multiple} and its variants such as~\cite{xiao2023direction,fernando2022mitigating,liu2021conflict,liu2023famo} tends to strike a max-min fairness among tasks, where the least-fortune tasks (i.e., with the lowest progress) are the most important. 
Nash-MTL~\cite{navon2022multi} aims to achieve proportional fairness among tasks by formulating the problem as a bargaining game, attaining a balanced solution that is not dominated by any single large gradient.

However, different applications may favor different types of task fairness, and there is currently no unified framework in MTL that allows for the incorporation of diverse fairness concepts beyond those previously mentioned. To fill this gap, we propose a novel fair  MTL framework, as well as efficient algorithms with performance guarantee. Our specific contributions are summarized below.

\begin{itemize}

    \item We first draw an important connection between MTL and fair resource allocation in communication networks~\cite{jain1984quantitative,kelly1997charging,mo2000fair,radunovic2007unified,srikant2013communication,ju2014optimal,liu2015optimal}, where we think of the common search direction $d$ shared by all tasks as a resource to minimize their losses, and the service quality is measure by the loss decrease after performing a gradient descent along $d$. Inspired by this connection, we model MTL as a utility maximization problem, where each task is associated with $\alpha$-fair utility function and different $\alpha$ yields different ideas of fairness including max-min, proportional, minimum potential delay fairness, etc. 
    
    \item We propose a novel algorithm named {\em \bf FairGrad} to solve the $\alpha$-fair MTL utility maximization problem. FairGrad is easy to implement, allows for a flexible selection of $\alpha$, and guarantees convergence to a Parato stationary point under mild assumptions. 
    
    \item Extensive experiments show that our FairGrad method can achieve state-of-the-art overall performance among gradient manipulation methods on 5 benchmarks in supervised learning and reinforcement learning with the number of tasks from 2 to 40.
    
    \item Finally, we incorporate our idea of $\alpha$-fairness into the loss functions of existing methods including Linear Scalarization, RLW, DWA, UW, MGDA, PCGrad, and CAGrad, and demonstrate that it can significantly improve their overall performance. 

\end{itemize}

\begin{figure*}[ht]
\vskip 0.2in
\begin{center}

    \subfigure[FairGrad (without fairness)]{
         \includegraphics[width=0.24\textwidth]{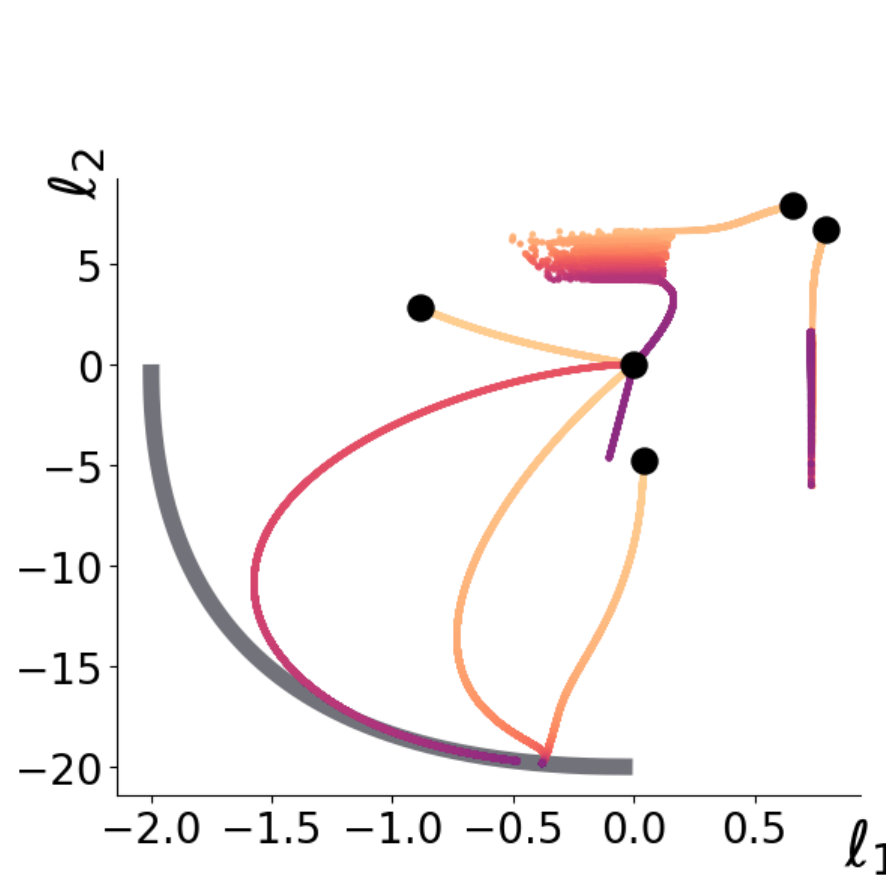}}
    \subfigure[FairGrad (proportionally fair)]{
         \includegraphics[width=0.24\textwidth]{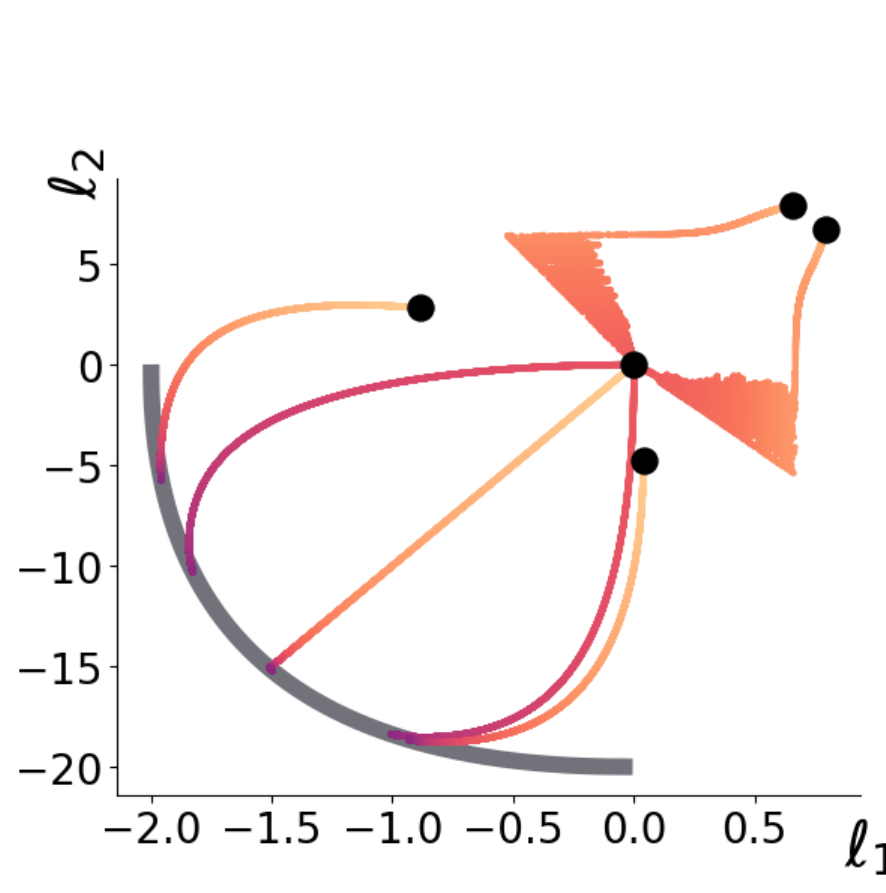}}
    \subfigure[FairGrad (MPD fair)]{
         \includegraphics[width=0.24\textwidth]{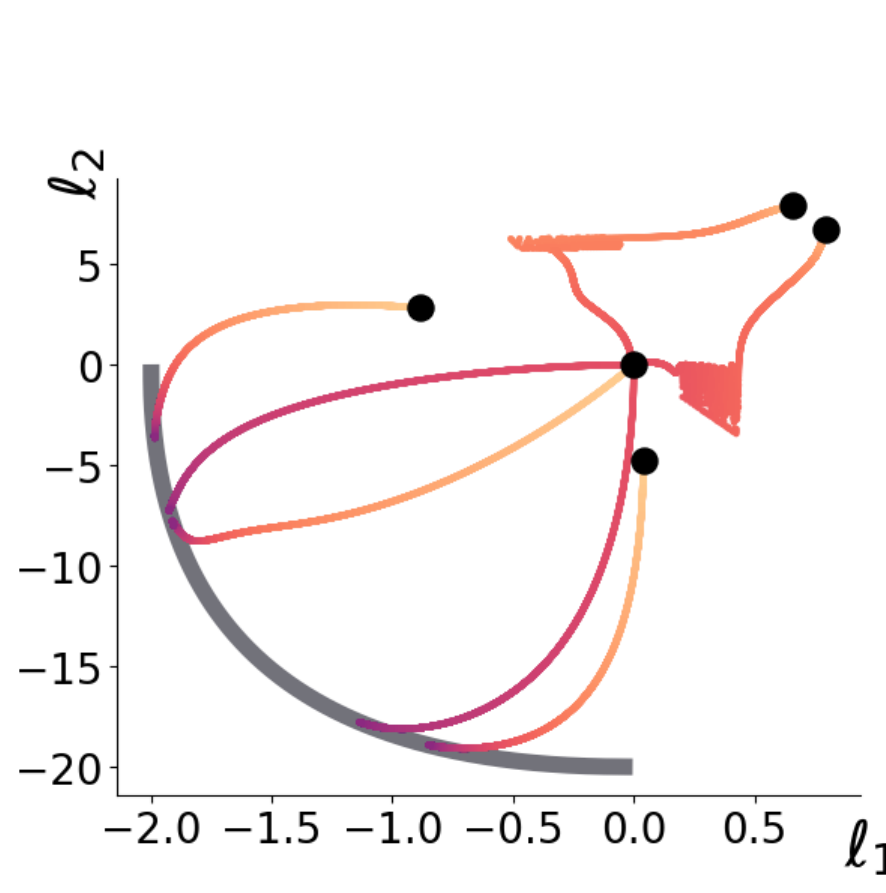}}
    \subfigure[FairGrad (max-min fair)]{
         \includegraphics[width=0.24\textwidth]{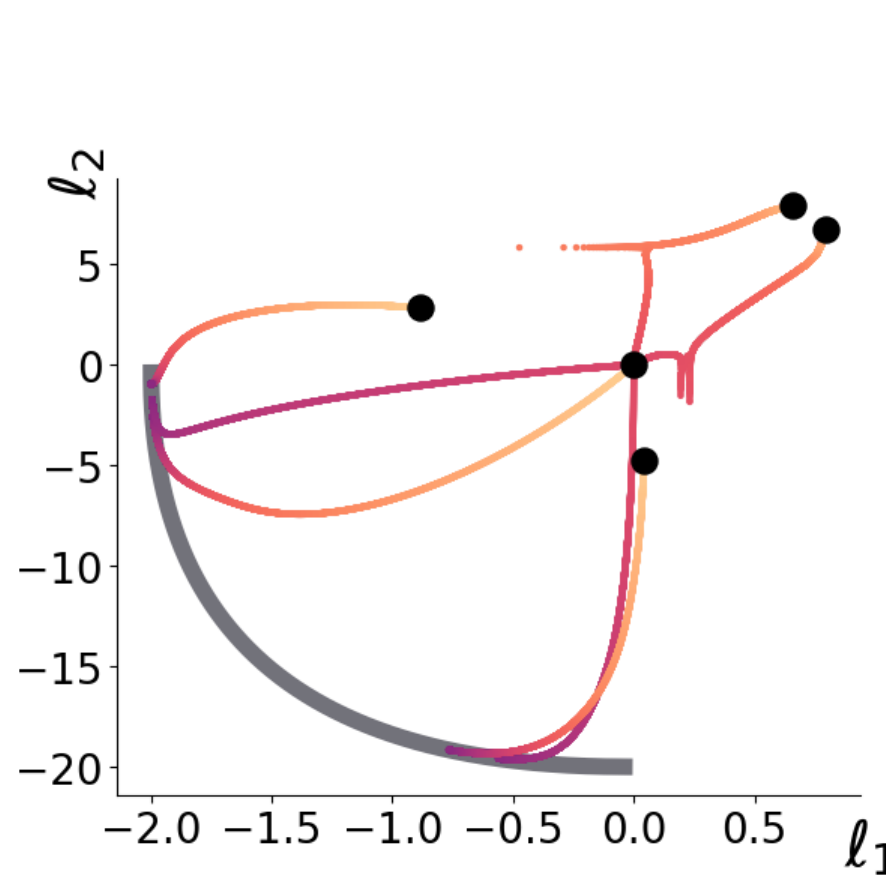}}
\caption{An illustrative two-task example from \cite{navon2022multi} to show the convergence of FairGrad to Pareto front from different initialization points (black dots $\bullet$). The optimization trajectories are colored from orange to purple. The bold gray line represents the Pareto front. The illustration showcases four 
fairness concepts (from left to right): simple average (i.e., Linear Scalarization (LS)), proportional fairness, minimum potential delay (MPD) fairness, and max-min fairness. It can be seen that LS is inclined towards the task $2$ with a larger gradient. FairGrad with proportional fairness resembles Nash-MTL~\cite{navon2022multi}, and can find more balanced solutions along the Pareto front.
MPD fairness aims to minimize the overall time for all tasks to converge, and shifts slightly more attention to some struggling tasks with smaller gradients. Max-min fairness emphasizes more on the less-fortune task with a smaller gradient magnitude. Also, observe that our FairGrad ensures the convergence to the Pareto front from all different initialization points.
}
\label{fig:toy_example}
\end{center}
\vskip -0.2in
\end{figure*}

\section{Related Work}

\textbf{Multi-Task Learning.} MTL has drawn significant attention both in theory and practice. One class of studies is designing sophisticated model architectures. These studies can be mainly divided into two categories, hard parameter sharing where task-specific layers are built on a common feature space \cite{liu2019end,kokkinos2017ubernet}, and soft parameter sharing which couples related parameters through certain constraints \cite{ruder2019latent,gao2020mtl}. Another line of research aims to capture the relationship among tasks to guide knowledge transfer effectively \cite{zhao2020efficient,ciliberto2017consistent}. Additionally, the magnitudes of losses for different tasks may vary, posing challenges to the optimization of MTL. 
A group of studies seeks to balance tasks through heuristic re-weighting rules such as task-dependent uncertainty \cite{kendall2018multi}, gradient magnitudes \cite{chen2018gradnorm}, and the rate of change of loss for each task \cite{liu2019end}. 

As MTL is one of the important applications of multi-objective optimization (MOO), several MOO-based gradient manipulation methods have been explored recently to address the challenge of conflicting gradients. \cite{desideri2012multiple} proposed MGDA, and show it guarantees the convergence to the Pareto front under certain assumptions. \cite{sener2018multi} cast MTL as a MOO problem and refined MGDA for optimization in the context of deep neural networks. \cite{yu2020gradient} determined the update by projecting a task's gradient onto the normal plane of other conflicting gradients. \cite{liu2021conflict} limited the update to a neighborhood of the average gradient. \cite{navon2022multi} considered finding the update as a bargaining game across all tasks. \cite{liu2023famo} searched for the update with the largest worst-case loss improvement rate to ensure that all tasks are optimized with approximately similar progress. 

Theoretically, \cite{hu2023revisiting} showed that Linear Scalarization cannot fully explore the Pareto front compared with MGDA-variant methods. \cite{zhou2022convergence}  
proposed a correlation-reduced stochastic gradient manipulation method to address the 
non-convergence issue of MGDA, CAGrad, and PCGrad in the stochastic setting. 
\cite{fernando2022mitigating} introduced a stochastic variant of MGDA with guaranteed convergence. \cite{xiao2023direction} proposed a simple and provable SGD-type method that benefits from direction-oriented improvements like \cite{liu2021conflict}. \cite{chen2023three} offered a framework for analyzing stochastic MOO algorithms, considering the trade-off among optimization, generalization, and conflict-avoidance.

\textbf{Fairness in Resource Allocation.} Fair resource allocation has been studied for decades in wireless communication~\cite{nandagopal2000achieving,eryilmaz2006joint,lan2010axiomatic,huaizhou2013fairness,noor2020survey,xu2021survey}, where limited resources such as power and communication bandwidth need to be fairly allocated to users of the networks. Various fairness criteria have been proposed to improve the service quality for all users without sacrificing the overall network throughput. For example, 
Jain's fairness index~\cite{jain1984quantitative} prefers all users to share the resources equally. Proportional fairness \cite{kelly1997charging} distributes resources proportional to user demands or priorities. Max-min fairness \cite{radunovic2007unified} attempts to protect the user who receives the least amount of resources by providing them with the maximum possible allocation. The $\alpha$-fairness framework was proposed to unify multiple fairness criteria, where different choices of $\alpha$ lead to different ideas of fairness \cite{mo2000fair,lan2010axiomatic}. Recent research has explored the application of fair resource allocation in federated learning~\cite{li2019fair,zhang2022proportional}. 
In this paper, we connect MTL with fair resource allocation and further propose an $\alpha$-fair utility maximization problem as well as an efficient algorithm to solve it.

\section{Preliminaries}

\subsection{Multiple Objectives and Pareto Concepts}

MTL involves multiple objective functions, denoted as  $L(\theta)=(l_1(\theta),\cdots,l_K(\theta))$, where $\theta$ represents model parameters, $
K$ is the number of tasks, and $l_i$ refers to the loss function of the $i$-th task. 

Given two points $\theta_1,\theta_2\in\mathbb{R}^m$, we say that $\theta_1$ \emph{dominates} $\theta_2$ if $l_i(\theta_1)\le l_i(\theta_2)$ for all $i\in[K]$ and $L(\theta_1)\neq L(\theta_2)$. A point is considered \emph{Pareto optimal} if it is not dominated by any other point. This means that no improvement can be made in any one objective without negatively affecting at least one other objective. The set of all Pareto optimal points form into the \emph{Pareto front}. A point $\theta\in\mathbb{R}^m$ is \emph{Pareto stationary} if $\min_{w\in\mathcal{W}} \|G(\theta)w\|=0$, where $G(\theta)=[g_1(\theta),\cdots,g_K(\theta)]\in\mathbb{R}^{m\times K}$ is the matrix with each column $g_i(\theta)$ denoting the gradient of $i$-th objective, and $\mathcal{W}$ is the probability simplex defined on $[K]$.  Pareto stationarity is a necessary condition for Pareto optimality.

\subsection{$
\alpha$-Fair Resource Allocation} \label{sec:fairRA_net}

In the context of fair resource allocation in communication networks with $K$ users, the goal is to properly allocate resources (e.g., channel bandwidth, transmission rate) to maximize the total user utility (e.g., throughput) under the link capacity constraints. A generic overall objective for the network is given by 
\begin{align}
\label{eq:alpha_fair_utility}
   \max_{x_1,...,x_K\in\mathcal{D}} \quad \sum_{i\in[K]}u(x_i):=\frac{x_i^{1-\alpha}}{1-\alpha},
\end{align}
where $\frac{x_i^{1-\alpha}}{1-\alpha}$ is the $\alpha$-fair function with  $\alpha\in[0,1)\cup(1,+\infty)$, and $x_i$ denotes  the packet transmission rate, and $\mathcal{D}$ denotes the convex link capacity constraints. Different $\alpha$ implements 
different ideas of fairness, as elaborated below. Let $x_i^*$ be the solution of the utility maximization problem. 

\textbf{Proportional Fairness}. This type of fairness is achieved when $\alpha\to 1$.
To see this, the user utility function then becomes $\log x_i$, and it can be shown (see \Cref{diff_alpha}) that    
    $\sum_i \frac{x_i-x_i^\star}{x_i^\star}\le 0$ for any $x_i\in\mathcal{D}$.
This inequality indicates that if the amount of resource assigned to one user is increased, then the sum of the proportional changes of all other users is non-positive and hence there is at least one other user with a \textbf{negative} proportional change. Thus, $x^*_i,i=1,...,K$ are called proportionally fair.

\textbf{Minimum Potential Delay Fairness}. When $\alpha=2$, the utility function is $-\frac{1}{x_i}$ and hence the overall objective is to minimize $\sum_i\frac{1}{x_i}$. Since $x_i$ is the transmission rate in networks, $\frac{1}{x_i}$ can be reviewed as the delay of transferring a file with unit size, and hence this case is called minimum potential delay fairness.

\textbf{Max-Min Fairness}. When $\alpha\to+\infty$, it is shown from \Cref{diff_alpha} that for any feasible allocation $x_i,i=1,...,K\in\mathcal{D}$, if $x_i>x_i^\star$ for some user $i$ then there exists another user $j$ such that $x_j^\star\le x_i^\star$ and $x_j<x_j^\star$, which further indicates that $ \min_i x_i^\star\ge\min_i x_i$. Thus, max-min fairness tends to protect the user who receives the least amount of resources by providing them with the maximum possible allocation.

\section{FairGrad: Fair Resource Allocation in MTL}
\label{method}

Inspired by the fair resource allocation over networks in \Cref{sec:fairRA_net}, we now provide an $\alpha$-fair framework for MTL. Let $d$ be the updating direction for all tasks within the ball $B_\epsilon$ centered at $0$ with a radius of $\epsilon$, and $g_i$ be the gradient for task $i$. Then, based on  the first-order Taylor approximation of $l_i(\theta)$, we have for a small stepsize $\eta$
\begin{align*}
    \frac{1}{\eta}[l_i(\theta)-l_i(\theta-\eta d)] \approx g_i^\top d,
\end{align*}
and hence $g_i^\top d$ can be regarded as the loss decreasing rate that plays a role similar to the transmission rate $x_i$ in communication networks. Towards this end, we define the $\alpha$-fair utility for each user $i$ as $\frac{(g_i^\top d)^{1-\alpha}}{1-\alpha}$, and hence the overall objective is to maximize the following total utilities of all tasks: 
\begin{align}
\label{eq:alpha_fair_mtl_obj}
\begin{split}
    &\max_{d\in B_\epsilon} \sum_{i\in [K]} \frac{(g_i^\top d)^{1-\alpha}}{1-\alpha}\\
    &\; \text{s.t.}\;\; \ g_i^\top d\ge0,
\end{split}
\end{align}
where $\alpha\in[0, 1)\cup(1,+\infty)$.
Note that our $\alpha$-fair framework takes the same spirit as Linear Scalarization (LS), Nash-MTL, and MGDA when we take $\alpha\to 0,1,\infty$, respectively, and provides the coverage over other fairness ideas. 

\subsection{Analogy}

\textbf{Utility}. For each user $i$ in a communication network, it usually holds that the larger the allocated transmission rate $x_i$, the higher the user’s level of satisfaction. However, the total link capacity in a communication network is constrained. 
For each task $i$ in MTL scenarios, the larger the loss decreasing rate $g_i^\top d$, the more optimized the task becomes. 
As we consider the update direction $d$ within a ball $B_\epsilon$ centered at 0 with a radius of $\epsilon$, the feasible update progress for each task is also constrained.
In a communication network, increasing the allocated transmission rate for one user may decrease the rate of other users. Similarly, conflicting gradients may occur in MTL.

\textbf{Capacity Constraint}. In the network resource allocation, the convex link capacity constraint refers to the constraints imposed on individual network links to ensure the packet transmission rate across each link does not exceed its capacity. It can be formulated as follows: $$\sum_{i\in L} x_i < C,$$ where $x_i\ge 0$, $C$ represents the capacity of the network link, and $L$ denotes the users who transmit their packets on this link.
In \cref{eq:alpha_fair_mtl_obj}, the loss decrease rate $g_i^\top d$ is modeled as a utility. The feasible update direction $d$ in the ball $B_\epsilon$. This is similar to the capacity constraint in network resource allocation because $d$ here cannot be arbitrarily large and then has capacity in MTL. In addition, the constraint on $d$ in our case is: $g_i^\top d\ge 0$ and $d\in B_\epsilon$ for all $i$, which turns out to be convex, as in network resource allocation where the capacity constraint is convex.

\subsection{Method}

We next take the following steps to solve the problem in \cref{eq:alpha_fair_mtl_obj}. First note that the objective function is non-decreasing with respect to any feasible $d$. Thus, if $d$ lies in the interior of $B_\epsilon$, then there must exist a point along the same direction but on the boundary of $B_\epsilon$, which achieves a larger overall utility. Thus, it can be concluded that the optimal $d^*$ lies on the boundary, and the gradient of the overall objective is aligned with $d^*$, i.e., 
\begin{align}\label{eq:alignment}
\sum_i g_i(g_i^\top d)^{-\alpha}=cd
\end{align}
for some constant $c>0$. Following Nash-MTL~\cite{navon2022multi}, we take $c=1$ for simplicity, and assume that the gradients of tasks are linearly independent when not at a Pareto stationary point $\theta$ such that $d$ can be represented as a linear combination of task gradients: $d=\sum_iw_ig_i$, where $w:=(w_1,...,w_K)^\top\in\mathbb{R}_{+}^K$ denotes the weights. Then, we obtain from \cref{eq:alignment} that $(g_i^\top d)^{-\alpha}=w_i$, which combined with $d=\sum_iw_ig_i$, implies that     
\begin{align}
\label{eq:alpha_fair_weight}
    G^\top Gw=w^{-1/\alpha},
\end{align}
where $\alpha\neq0$ and the power $-1/\alpha$ is applied elementwisely. It is evident that we have $w_i=1$ for all $i\in[K]$ when $\alpha=0$. Differently from Nash-MTL that approximates the solution using a sequence of convex
optimization problems, we treat \cref{eq:alpha_fair_weight} as a simple constrained nonlinear least square problem 
\begin{align*}
\begin{split}
    &\min_w \sum_i f(w)_i^2\\
    &\; \text{s.t.}\;\;  f(w)=G^\top Gw - w^{-1/\alpha}\ \ w\in\mathbb{R}_{+}^K,\\
\end{split}
\end{align*}
which is solved by  
 {\em scipy.optimize.least\_squares} efficiently. The complete procedure of our algorithm is summarized in \Cref{alg:alpha_fair_mtl}.

 \begin{algorithm}[tb]
   \caption{FairGrad for MTL}
   \label{alg:alpha_fair_mtl}
\begin{algorithmic}[1]
   \STATE {\bfseries Input:} Model parameters $\theta_0$, $\alpha$, learning rate $\{\eta_t\}$
   \FOR{$t=1$ {\bfseries to} $T-1$}
   \STATE Compute gradients $G(\theta_t)=[g_1(\theta_t),\cdots,g_K(\theta_t)]$
   \STATE Solve \cref{eq:alpha_fair_weight} to obtain $w_t$
   \STATE Compute $d_t=G(\theta_t)w_t$
   \STATE Update the parameters $\theta_{t+1}=\theta_t-\eta d_t$
   \ENDFOR
\end{algorithmic}
\end{algorithm}

\section{Empirical Results}

We first use a toy example to elaborate how FairGrad balances the tasks by incorporating different fairness criteria.
Then we conduct extensive experiments under both supervised learning and reinforcement learning settings to demonstrate the effectiveness of our proposed method. Full experimental details and more empirical studies can be found in \Cref{app:experiments}.

\subsection{Toy Example}

We adopt the 2-task toy example introduced in \cite{navon2022multi}, where the objectives of the 2 tasks, denoted as $L_1$ and $L_2$, have different scales. More details are provided in \Cref{app:toy}. We select 5 starting points and illustrate the optimization trajectories of FairGrad with different fairness criteria in \Cref{fig:toy_example}. 

Obviously, without fairness (Linear Scalarization), the algorithm may not converge to a Pareto stationary point. However, in other cases where fairness is involved, the algorithm can converge. 
Furthermore, in the experiment setting, objective $L_2$ exhibits a larger scale than objective $L_1$, resulting in a larger gradient magnitude. If there is no fairness, task 2 will dominate the optimization process. When the algorithm converges, it will always converge to a stationary point where $L_2$ is smaller than $L_1$, as shown in \Cref{fig:toy_example}. On the other hand, with max-min fairness, the least fortunate task will be prioritized. The algorithm tends to converge to a stationary point with a smaller $L_1$. Proportional fairness and MPD fairness will lead to a more balanced solution.

\begin{table}[t]
\caption{Results on CelebA (40-task) and QM9 (11-task) datasets. Each experiment is repeated 3 times with different random seeds and the average is reported.}
\label{tab:celeba_qm9}
\vskip 0.15in
\begin{center}
\begin{small}
\begin{sc}
\begin{adjustbox}{max width=0.48\textwidth}
\begin{tabular}{lccccr}
\toprule
\multirow{2}*{Method} & \multicolumn{2}{c}{CelebA} & \multicolumn{2}{c}{QM9} \\
\cmidrule(lr){2-3}\cmidrule(lr){4-5}
& MR $\downarrow$ & $\Delta m\%\downarrow$ & MR $\downarrow$ & $\Delta m\%\downarrow$ & \\
\midrule
LS & 6.53 & 4.15 & 8.18 & 177.6 \\
SI & 8.00 & 7.20 & 4.82 & 77.8 \\
RLW & 5.40 & 1.46 & 9.55 & 203.8 \\
DWA & 7.23 & 3.20 & 7.82 & 175.3 \\
UW & 6.00 & 3.23 & 6.18 & 108.0 \\
MGDA & 11.05 & 14.85 & 7.73 & 120.5 \\
PCGrad & 6.98 & 3.17 & 6.36 & 125.7 \\
CAGrad & 6.53 & 2.48 & 7.18 & 112.8 \\
IMTL-G & \textbf{4.95} & 0.84 & 6.09 & 77.2 \\
Nash-MTL & 5.38 & 2.84 & \textbf{3.64} & 62.0 \\
FAMO & 5.03 & 1.21 & 4.73 & 58.5 \\
\midrule
FairGrad & \textbf{4.95} & \textbf{0.37} & 3.82 & \textbf{57.9} \\
\bottomrule
\end{tabular}
\end{adjustbox}
\end{sc}
\end{small}
\end{center}
\vskip -0.1in
\end{table}

\begin{table*}[t]
\caption{Results on NYU-v2 (3-task) dataset. Each experiment is repeated 3 times with different random seeds and the average is reported.}
\label{tab:nyuv2}
\vskip 0.15in
\begin{center}
\begin{small}
\begin{sc}
\begin{adjustbox}{max width=\textwidth}
  \begin{tabular}{llllllllllll}
    \toprule
    \multirow{3}*{Method} & \multicolumn{2}{c}{Segmentation} & \multicolumn{2}{c}{Depth} & \multicolumn{5}{c}{Surface Normal} & \multirow{3}*{MR $\downarrow$} & \multirow{3}*{$\Delta m\%\downarrow$} \\
    \cmidrule(lr){2-3}\cmidrule(lr){4-5}\cmidrule(lr){6-10}
    & \multirow{2}*{mIoU $\uparrow$} & \multirow{2}*{Pix Acc $\uparrow$} & \multirow{2}*{Abs Err $\downarrow$} & \multirow{2}*{Rel Err $\downarrow$} & \multicolumn{2}{c}{Angle Distance $\downarrow$} & \multicolumn{3}{c}{Within $t^\circ$ $\uparrow$} & \\
    \cmidrule(lr){6-7}\cmidrule(lr){8-10}
    & & & & & Mean & Median & 11.25 & 22.5 & 30 & \\
    \midrule
    STL & 38.30 & 63.76 & 0.6754 & 0.2780 & 25.01 & 19.21 & 30.14 & 57.20 & 69.15 & \\
    \midrule
    LS & 39.29 & 65.33 & 0.5493 & 0.2263 & 28.15 & 23.96 & 22.09 & 47.50 & 61.08 & 10.67 & 5.59  \\
    SI & 38.45 & 64.27 & 0.5354 & 0.2201 & 27.60 & 23.37 & 22.53 & 48.57 & 62.32 & 9.44 & 4.39 \\
    RLW & 37.17 & 63.77 & 0.5759 & 0.2410 & 28.27 & 24.18 & 22.26 & 47.05 & 60.62 & 13.11 & 7.78 \\
    DWA & 39.11 & 65.31 & 0.5510 & 0.2285 & 27.61 & 23.18 & 24.17 & 50.18 & 62.39 & 9.44 & 3.57 \\
    UW & 36.87 & 63.17 & 0.5446 & 0.2260 & 27.04 & 22.61 & 23.54 & 49.05 & 63.65 & 9.22 & 4.05 \\
    MGDA & 30.47 & 59.90 & 0.6070 & 0.2555 & 24.88 & \textbf{19.45} & 29.18 & \textbf{56.88} & \textbf{69.36} & 7.11 & 1.38 \\
    PCGrad & 38.06 & 64.64 & 0.5550 & 0.2325 & 27.41 & 22.80 & 23.86 & 49.83 & 63.14 & 9.78 & 3.97 \\
    GradDrop & 39.39 & 65.12 & 0.5455 & 0.2279 & 27.48 & 22.96 & 23.38 & 49.44 & 62.87 & 8.78 & 3.58 \\
    CAGrad & 39.79 & 65.49 & 0.5486 & 0.2250 & 26.31 & 21.58 & 25.61 & 52.36 & 65.58 & 5.78 & 0.20 \\
    IMTL-G & 39.35 & 65.60 & 0.5426 & 0.2256 & 26.02 & 21.19 & 26.20 & 53.13 & 66.24 & 5.11 & -0.76 \\
    MoCo & \textbf{40.30} & \textbf{66.07} & 0.5575 & \textbf{0.2135} & 26.67 & 21.83 & 25.61 & 51.78 & 64.85 & 5.44 & 0.16 \\
    Nash-MTL & 40.13 & 65.93 & \textbf{0.5261} & 0.2171 & 25.26 & 20.08 & 28.40 & 55.47 & 68.15 & 3.11 & -4.04 \\
    FAMO & 38.88 & 64.90 & 0.5474 & 0.2194 & 25.06 & 19.57 & 29.21 & 56.61 & 68.98 & 4.44 & -4.10 \\
    \midrule
    FairGrad & 39.74 & 66.01 & 0.5377 & 0.2236 & \textbf{24.84} & 19.60 & \textbf{29.26} & 56.58 & 69.16 & \textbf{2.67} & \textbf{-4.66} \\
    \bottomrule
  \end{tabular}
  \end{adjustbox}
\end{sc}
\end{small}
\end{center}
\vskip -0.1in
\end{table*}

\subsection{Supervised Learning}
\label{exp:supervised}
We evaluate the performance of our method in three different supervised learning scenarios described as follows.

\textbf{Image-Level Classification.} CelebA \cite{liu2015faceattributes} is a large-scale face attributes dataset, containing over 200K celebrity images. Each image is annotated with 40 attributes, such as smiling, wavy hair, mustache, etc. We can consider the dataset as an image-level 40-task MTL classification problem, with each task predicting the presence of a specific attribute. This setting assesses the capability of MTL methods in handling a large number of tasks. We follow the experiment setup in \cite{liu2023famo}. We employ a network containing a 9-layer convolutional neural network (CNN) as the backbone and a specific linear layer for each task. We train our method for 15 epochs, using Adam optimizer with learning rate 3e-4. The batch size is 256.

\textbf{Regression.} QM9 \cite{ramakrishnan2014quantum} is a widely-used benchmark in graph neural networks. It comprises over 130k organic molecules, which are organized as graphs with annotated node and edge features. The goal of predicting 11 properties with different measurement scales is to see if MTL methods can effectively balance the variations present across these tasks. Following \cite{navon2022multi,liu2023famo}, we use the example provided in Pytorch Geometric \cite{fey2019fast}, and use 110k molecules for training, 10k for validation, and the rest 10k for testing. We train our method for 300 epochs with a batch size of 120. The initial learning rate is 1e-3, and a scheduler is used to reduce the learning rate once the improvement of validation stagnates.

\textbf{Dense Prediction.} NYU-v2 \cite{silberman2012indoor} contains 1449 densely annotated images that have been collected from video sequences of various indoor scenes. It involves one pixel-level classification task and two pixel-level regression tasks, which correspond to 13-class semantic segmentation, depth estimation, and surface normal prediction, respectively. Similarly, Cityscapes \cite{Cordts2016Cityscapes} contains 5000 street-scene images with two tasks: 7-class semantic segmentation and depth estimation. This scenario evaluates the effectiveness of MTL methods in tackling complex situations. We follow \cite{liu2021conflict,navon2022multi,liu2023famo} and adopt the backbone of MTAN \cite{liu2019end}, which adds task-specific attention modules on SegNet \cite{badrinarayanan2017segnet}. We train our method for 200 epochs with batch size 2 for NYU-v2 and 8 for Cityscapes. The learning rate is 1e-4 for the first 100 epochs, then decayed by half for the rest.

\textbf{Evaluation.} For image-level classification and regression, we compare our FairGrad with Linear Scalarization (LS) which minimizes the sum of task losses, Scale-Invariant (SI) which minimizes the sum of logarithmic losses, Random Loss Weighting (RLW) \cite{lin2021reasonable}, Dynamic Weigh Average (DWA) \cite{liu2019end}, Uncertainty weighting (UW) \cite{kendall2018multi}, MGDA \cite{sener2018multi}, PCGrad \cite{yu2020gradient}, CAGrad \cite{liu2021conflict}, IMTL-G \cite{liu2020towards}, Nash-MTL \cite{navon2022multi}, and FAMO \cite{liu2023famo}. For dense prediction, we also compare with GradDrop \cite{chen2020just}, and MoCo \cite{fernando2022mitigating}. We consider two metrics to represent the overall performance of the MTL method $m$. \textbf{(1)} $\mathbf{\Delta m\%}$, the average per-task performance drop against the single-task (STL) baseline $b$: $$\Delta m\%=\frac{1}{K}\sum_{i=1}^K(-1)^{\delta_k}(M_{m,k}-M_{b,k})/M_{b,k}\times100,$$ where $M_{b,k}$ denotes the value of metric $M_k$ from baseline $b$, $M_{m,k}$ denotes the value of metric $M_k$ from the compared method $m$, and $\delta_k=1$ if metric $M_k$ prefers a higher value. \textbf{(2)} \textbf{Mean Rank (MR)}, the average rank of each metric across tasks. 

\begin{table*}[t]
\caption{Results on Cityscapes (2-task) dataset. Each experiment is repeated 3 times with different random seeds and the average is reported.}
\label{tab:cityscapes}
\vskip 0.15in
\begin{center}
\begin{small}
\begin{sc}
\begin{adjustbox}{max width=\textwidth}
  \begin{tabular}{lllllll}
    \toprule
    \multirow{2}*{Method} & \multicolumn{2}{c}{Segmentation} & \multicolumn{2}{c}{Depth} & 
    \multirow{2}*{MR $\downarrow$} &
    \multirow{2}*{$\Delta m\%\downarrow$} \\
    \cmidrule(lr){2-3}\cmidrule(lr){4-5}
    & mIoU $\uparrow$ & Pix Acc $\uparrow$ & Abs Err $\downarrow$ & Rel Err $\downarrow$ & \\
    \midrule
    STL & 74.01 & 93.16 & 0.0125 & 27.77 & \\
    \midrule
    LS & 75.18 & 93.49 & 0.0155 & 46.77 & 8.50 & 22.60  \\
    SI & 70.95 & 91.73 & 0.0161 & 33.83 & 10.50 & 14.11 \\
    RLW & 74.57 & 93.41 & 0.0158 & 47.79 & 10.75 & 24.38 \\
    DWA & 75.24 & 93.52 & 0.0160 & 44.37 & 8.50 & 21.45 \\
    UW & 72.02 & 92.85 & 0.0140 & \textbf{30.13} & 6.75 & 5.89 \\
    MGDA & 68.84 & 91.54 & 0.0309 & 33.50 & 11.00 & 44.14  \\
    PCGrad & 75.13 & 93.48 & 0.0154 & 42.07 & 8.50 & 18.29  \\
    GradDrop & 75.27 & 93.53 & 0.0157 & 47.54 & 8.00 & 23.73  \\
    CAGrad & 75.16 & 93.48 & 0.0141 & 37.60 & 7.00 & 11.64  \\
    IMTL-G & 75.33 & 93.49 & 0.0135 & 38.41 & 5.50 & 11.10 \\
    MoCo & 75.42 & 93.55 & 0.0149 & 34.19 & 4.50 & 9.90 \\
    Nash-MTL & 75.41 & 93.66 & \textbf{0.0129} & 35.02 & 3.25 & 6.82 \\
    FAMO & 74.54 & 93.29 & 0.0145 & 32.59 & 7.25 & 8.13 \\
    \midrule
    FairGrad & \textbf{75.72} & \textbf{93.68} & 0.0134 & 32.25 & \textbf{1.50} & \textbf{5.18} \\
    \bottomrule
  \end{tabular}
  \end{adjustbox}
\end{sc}
\end{small}
\end{center}
\vskip -0.1in
\end{table*}

\textbf{Results.} The experiment results are shown in \cref{tab:celeba_qm9}, \cref{tab:nyuv2}, and \cref{tab:cityscapes}. Each experiment is repeated 3 times with different random seeds and the average is computed. 
It can be seen from \cref{tab:celeba_qm9} that the proposed FairGrad outperforms existing methods on the CelebA dataset with 40 tasks, indicating that it performs effectively when faced with a substantial number of tasks. \cref{tab:celeba_qm9} shows that FairGrad also achieves the best overall performance drop $\Delta m\%$ on the QM9 dataset, while attaining a mean rank of 3.82 comparable to the best 3.64 of Nash-MTL. 
In addition, \cref{tab:nyuv2} and \cref{tab:cityscapes} show that FairGrad outperforms all the baselines on the NYU-v2 and Cityscapes datasets w.r.t.~MR and $\Delta m\%$, demonstrating its effectiveness in learning from scene understanding scenarios. 

Furthermore, there are some other interesting findings from the results presented in \cref{tab:nyuv2}. LS performs poorly in the surface normal prediction (SNP) task compared to the other two tasks. This is because LS does not take the fairness among tasks into consideration, and hence the gradient of the SNP task is dominated by the others. On the contrary, MGDA \cite{sener2018multi} obtains the best performance in the SNP task among all three tasks by enforcing the max-min fairness. Meanwhile, the performance of Nash-MTL \cite{navon2022multi}, which embodies proportional fairness, is more balanced across all tasks. As a comparison,  our FairGrad can find a more balanced solution than LS and MGDA, while placing greater emphasis on the challenging SNP tasks than Nash-MTL.

\subsection{Reinforcement Learning}
\label{exp:reinforcement}
We further evaluate our method on the MT10, a benchmark including 10 robotic manipulation tasks from the MetaWorld environment \cite{yu2020meta}, where the objective is to learn one policy that generalizes to different tasks such as pick and place, open door, etc. We follow \cite{liu2021conflict,navon2022multi,liu2023famo} and adopt Soft Actor-Critic (SAC) \cite{haarnoja2018soft} as the underlying algorithm. We implement with MTRL codebase \cite{Sodhani2021MTRL} and train our method for 2 million steps with a batch size of 1280.

\begin{table}[t]
\caption{Results on MT10 benchmark. Average over 10 random seeds. Nash-MTL$^\star$ denotes the result reported in the original paper \cite{navon2022multi}. While Nash-MTL (reproduced) denotes the reproduced result in \cite{liu2023famo}.}
\label{tab:mt10}
\vskip 0.15in
\begin{center}
\begin{small}
\begin{sc}
\begin{adjustbox}{max width=\textwidth}
  \begin{tabular}{ll}
    \toprule
    \multirow{2}*{Method} & success rate\\
    & (mean ± stderr) \\
    \midrule
    STL & 0.90 $\pm$ 0.03\\
    \midrule
    MTL SAC & 0.49 $\pm$ 0.07 \\
    MTL SAC + TE & 0.54 $\pm$ 0.05 \\
    MH SAC & 0.61 $\pm$ 0.04 \\
    PCGrad & 0.72 $\pm$ 0.02 \\
    CAGrad & 0.83 $\pm$ 0.05 \\
    MoCo & 0.75 $\pm$ 0.05 \\
    Nash-MTL$^\star$ & \textbf{0.91} $\pm$ 0.03 \\
    Nash-MTL (reproduced) & 0.80 $\pm$ 0.13 \\
    FAMO & 0.83 $\pm$ 0.05 \\
    \midrule
    FairGrad & 0.84 $\pm$ 0.07\\
    \bottomrule
  \end{tabular}
  \end{adjustbox}
\end{sc}
\end{small}
\end{center}
\vskip -0.1in
\end{table}

\textbf{Evaluation.} We compare our FairGrad with Multi-task SAC (MTL SAC) \cite{yu2020meta}, Multi-task SAC with task encoder (MTL SAC + TE) \cite{yu2020meta}, Multi-headed SAC (MH SAC) \cite{yu2020meta}, PCGrad \cite{yu2020gradient}, CAGrad \cite{liu2021conflict}, MoCo \cite{fernando2022mitigating}, Nash-MTL \cite{navon2022multi}, and FAMO \cite{liu2023famo}. 

\textbf{Results.} The results are shown in \cref{tab:mt10}. Each method is evaluated once every 10,000 steps, and the best average success rate over 10 random seeds throughout the entire training course is reported. We could not reproduce the MTRL result in the original paper of Nash-MTL exactly, and hence we adopt the reproduced result of Nash-MTL in \cite{liu2023famo}. 
It is evident that our method performs competitively when compared to other methods.

\subsection{Effect of Different Fairness Criteria}\label{exp:fairness}

We investigate the effect of different fairness criteria on NYU-v2 and Cityscapes datasets by setting $\alpha\to[1,2,5,10]$, which corresponds to the proportional fairness, minimum potential delay fairness, and approximate max-min fairness. The results are presented in \cref{tab:fairness_effect}. The results show that different fairness criteria prioritize different tasks, and thus lead to different overall performance. In particular, the minimum potential delay fairness with $\alpha=2$ achieves the best $\Delta m\%$ among all fairness criteria. 

Also note that although the best $\Delta m\%$ reported in \cref{tab:fairness_effect} is better than that reported in \cref{tab:nyuv2} and \cref{tab:cityscapes}, their results w.r.t.~MR are worse than those in \cref{tab:nyuv2} and \cref{tab:cityscapes}. This is because an improved $\Delta m\%$ may result in a lower rank for certain tasks, causing a significant degradation in the average rank. See \cref{app:fairness_effect} for more details.

\begin{table}[H]
\caption{$\Delta m\%$ of different fairness criteria on NYU-v2 (3-task) and Cityscapes (2-task) datasets.}
\label{tab:fairness_effect}
\begin{center}
\begin{small}
\begin{sc}
\begin{adjustbox}{max width=\textwidth}
  \begin{tabular}{lll}
    \toprule
    Method & NYU-v2 & Cityscapes \\
    \midrule
    FairGrad ($\alpha=1$) & -2.79 & 6.73 \\
    FairGrad ($\alpha=2$) & \textbf{-4.96} & \textbf{3.90} \\
    FairGrad ($\alpha=5$) & -3.03 & 6.87 \\
    FairGrad ($\alpha=10$) & -1.00 & 10.54 \\
    \bottomrule
  \end{tabular}
  \end{adjustbox}
\end{sc}
\end{small}
\end{center}
\vskip -0.1in
\end{table}

\subsection{Discussion on Practical Implementation}
\label{exp:practical}

\textbf{Supervised Learning.} For experiments on QM9, NYU-v2, and Cityscapes, we implement our method based on the codes released by \cite{navon2022multi}. For experiments on CelebA, our implementation is based on the codes provided by \cite{liu2023famo}, consistent with all the baselines presented in \cref{tab:celeba_qm9}.

\textbf{Reinforcement Learning.} We find it time-consuming to solve the constrained nonlinear least square problem discussed in \cref{method} under the reinforcement learning setting. Therefore, we use SGD to approximately solve the problem and accelerate the training process. Specifically, we use SGD optimizer with a learning rate of 0.1, momentum of 0.5, and train 20 epochs.

\textbf{Choice of $\alpha$.} We first search with $\alpha\in[1,2,5,10]$ and evaluate which choice is better. Then we narrow down the search space and continue to execute a grid search with a step size of 0.1 until we determine an appropriate value. 

In practice, the choice depends on the specific needs or preferences. If there are no requirements for fairness and tasks with larger gradients are allowed to finish first, we can simply set $\alpha=0$ to allow for quick training. If tasks with struggling progress are prioritized (e.g., in some meta-learning setups, some harder-to-train tasks may play more important roles in deciding final test accuracy), then the max-min fairness (with a larger $\alpha$) is desired. From our observation, if aiming to achieve the most balanced overall performance, MPD fairness with $\alpha=2$ is preferable. After fairness criteria are selected, some slight finetuning on $\alpha$ can also be conducted to further improve the overall accuracy.

\begin{table*}[t]
\caption{Results of $\alpha$-fair loss  transformation on NYU-v2 (3-task) dataset. Each experiment is repeated 3 times with different random seeds and the average is reported. We simply choose $\alpha=0.5$.}
\label{tab:additional_nyuv2}
\vskip 0.15in
\begin{center}
\begin{small}
\begin{sc}
\begin{adjustbox}{max width=\textwidth}
  \begin{tabular}{lllllllllll}
    \toprule
    \multirow{3}*{Method} & \multicolumn{2}{c}{Segmentation} & \multicolumn{2}{c}{Depth} & \multicolumn{5}{c}{Surface Normal} & \multirow{3}*{$\Delta m\%\downarrow$} \\
    \cmidrule(lr){2-3}\cmidrule(lr){4-5}\cmidrule(lr){6-10}
    & \multirow{2}*{mIoU $\uparrow$} & \multirow{2}*{Pix Acc $\uparrow$} & \multirow{2}*{Abs Err $\downarrow$} & \multirow{2}*{Rel Err $\downarrow$} & \multicolumn{2}{c}{Angle Distance $\downarrow$} & \multicolumn{3}{c}{Within $t^\circ$ $\uparrow$} & \\
    \cmidrule(lr){6-7}\cmidrule(lr){8-10}
    & & & & & Mean & Median & 11.25 & 22.5 & 30 & \\
    \midrule
    LS & 39.29 & 65.33 & 0.5493 & 0.2263 & 28.15 & 23.96 & 22.09 & 47.50 & 61.08 & 5.59  \\
    Fair-LS & 38.64 & 64.96 & 0.5422 & 0.2255 & 27.14 & 22.64 & 24.05 & 50.14 & 63.53 & \textbf{2.85}  \\
    \midrule
    RLW & 37.17 & 63.77 & 0.5759 & 0.2410 & 28.27 & 24.18 & 22.26 & 47.05 & 60.62 & 7.78 \\
    Fair-RLW & 37.29 & 63.58 & 0.5481 & 0.2263 & 27.67 & 23.33 & 23.38 & 48.72 & 62.12 & \textbf{5.00} \\
    \midrule
    DWA & 39.11 & 65.31 & 0.5510 & 0.2285 & 27.61 & 23.18 & 24.17 & 50.18 & 62.39 & 3.57 \\
    Fair-DWA & 39.03 & 65.18 & 0.5404 & 0.2266 & 27.20 & 22.63 & 24.31 & 50.14 & 63.45 & \textbf{2.65} \\
    \midrule
    UW & 36.87 & 63.17 & 0.5446 & 0.2260 & 27.04 & 22.61 & 23.54 & 49.05 & 63.65 & 4.05 \\
    Fair-UW & 38.51 & 64.56 & 0.5423 & 0.2274 & 27.23 & 22.92 & 23.62 & 49.52 & 63.23 & \textbf{3.56} \\
    \midrule
    MGDA & 30.47 & 59.90 & 0.6070 & 0.2555 & 24.88 & 19.45 & 29.18 & 56.88 & 69.36 & 1.38 \\
    Fair-MGDA & 35.91 & 63.19 & 0.5646 & 0.2260 & 24.75 & 19.24 & 30.04 & 57.30 & 69.55 & \textbf{-3.26} \\
    \midrule
    PCGrad & 38.06 & 64.64 & 0.5550 & 0.2325 & 27.41 & 22.80 & 23.86 & 49.83 & 63.14 & 3.97 \\
    Fair-PCGrad & 39.26 & 65.08 & 0.5257 & 0.2177 & 26.88 & 22.26 & 24.74 & 50.85 & 64.18 & \textbf{1.23} \\
    \midrule
    CAGrad & 39.79 & 65.49 & 0.5486 & 0.2250 & 26.31 & 21.58 & 25.61 & 52.36 & 65.58  & 0.20 \\
    Fair-CAGrad & 39.32 & 65.36 & 0.5290 & 0.2221 & 25.50 & 20.32 & 28.06 & 54.94 & 67.65  & \textbf{-2.91} \\
    \bottomrule
  \end{tabular}
  \end{adjustbox}
\end{sc}
\end{small}
\end{center}
\vskip -0.1in
\end{table*}

\section{Applying $\alpha$-Fairness to Existing Methods}
In MTL, tasks often exhibit variations in difficulty, resulting in losses that may vary in scale. Since the idea of $\alpha$-fairness provides a framework unifying different fairness criteria, we argue that it can be directly applied in many MTL methods to mitigate the problem of varying loss scales by replacing the task losses $(l_1,\cdots,l_K)$ with
\begin{align}\label{alpha_trans}
    \Big(\frac{l_1^{1-\alpha}}{1-\alpha},\cdots,\frac{l_K^{1-\alpha}}{1-\alpha}\Big),
\end{align}
where $\alpha\in(-\infty,1)$ and $i\in[K]$. Note that the meaning of $\alpha$ here differs from that used in \cref{method}. FairGrad aims to address the issue of varying loss decreasing rates, while applying $\alpha$-fairness to existing methods tries to deal with the issue of varying loss scales. Although the ideas of $\alpha$-fairness are the same, the goals are different.

Here we omit the model parameter $\theta$ for simplicity. It can be observed that the gradient changes from $g_i$ to $g_i/l_i^\alpha$, where $\alpha$ controls the emphasis placed on tasks with different levels of difficulty. Take the example of simply summing $\alpha$-fair  losses of all tasks
\begin{align*}
    \min \sum_{i\in[K]} \frac{l_i^{1-\alpha}}{1-\alpha}.
\end{align*}
If we choose $\alpha=0$, the objective is reduced to Linear Scalarization (LS) which minimizes the sum of all losses. If $\alpha\to1$, the objective tends to minimize the sum of the logarithmic losses, which shares similarity with Scale-Invariant (SI). If $\alpha\to-\infty$, the objective exhibits the notion of the minimax fairness \cite{radunovic2007unified}, which aims to minimize the maximum loss among all tasks. 

\begin{proposition}
\label{prop:alpha_fair_loss}
    The Pareto front of the $\alpha$-fair loss functions in \cref{alpha_trans} is the same as that of original loss functions $(l_1,...,l_K)$.
\end{proposition}
According to \cref{prop:alpha_fair_loss},
transforming each $l_i$ to its $\alpha$-fair counterpart does not change the Pareto front, and allows us to find an improved solution along this front under a proper selection of fairness. 

We then apply this $\alpha$-fair loss transformation to a series of MTL methods including LS, RLW \cite{lin2021reasonable}, DWA \cite{liu2019end}, UW \cite{kendall2018multi}, MGDA \cite{sener2018multi}, PCGrad \cite{yu2020gradient}, CAGrad \cite{liu2021conflict}, and test the performance on NYU-v2 and Cityscapes datasets. We simply choose $\alpha=0.5$ for all the experiments. Other experiment settings remain the same with \cref{exp:supervised}. The results presented in \cref{tab:additional_nyuv2} and \cref{tab:additional_cityscapes} clearly demonstrate that the $\alpha$-fair loss transformation improves the performance of these MTL methods via a large margin. 

Additionally, we also test the applicability of $\alpha$-fair loss transformation to FairGrad. It can be seen from \cref{tab:additional_cityscapes} that compared to other MTL methods, applying this transformation to FairGrad provides only a marginal improvement. This shows that FairGrad can mitigate the issue of varying loss scales by incorporating fairness-based utility functions.

\begin{table}[t]
\caption{Results of $\alpha$-fair loss transformation on Cityscapes (2-task) dataset. We simply choose $\alpha=0.5$.}
\label{tab:additional_cityscapes}
\vskip 0.15in
\begin{center}
\begin{small}
\begin{sc}
\begin{adjustbox}{max width=0.48\textwidth}
  \begin{tabular}{llllll}
    \toprule
    \multirow{2}*{Method} & \multicolumn{2}{c}{Segmentation} & \multicolumn{2}{c}{Depth} &
    \multirow{2}*{$\Delta m\%\downarrow$} \\
    \cmidrule(lr){2-3}\cmidrule(lr){4-5}
    & mIoU $\uparrow$ & Pix Acc $\uparrow$ & Abs Err $\downarrow$ & Rel Err $\downarrow$ & \\
    \midrule
    LS & 75.18 & 93.49 & 0.0155 & 46.77  & 22.60 \\
    Fair-LS & 74.91 & 93.48 & 0.0137 & 37.51 & \textbf{10.86} \\
    \midrule
    RLW & 74.57 & 93.41 & 0.0158 & 47.79 & 24.38 \\
    Fair-RLW & 74.32 & 93.36 & 0.0140 & 37.46 & \textbf{11.64} \\
    \midrule
    DWA & 75.24 & 93.52 & 0.0160 & 44.37  & 21.45 \\
    Fair-DWA & 75.06 & 93.46 & 0.0147 & 35.34 & \textbf{10.74} \\
    \midrule
    MGDA & 68.84 & 91.54 & 0.0309 & 33.50 & 44.14 \\
    Fair-MGDA & 74.45 & 93.50 & 0.0131 & 37.64 & \textbf{9.91} \\
    \midrule
    PCGrad & 75.13 & 93.48 & 0.0154 & 42.07 & 18.29 \\
    Fair-PCGrad & 75.25 & 93.51 & 0.0140 & 37.00 & \textbf{10.71} \\
    \midrule
    CAGrad & 75.16 & 93.48 & 0.0141 & 37.60 & 11.64  \\
    Fair-CAGrad & 74.74 & 93.39 & 0.0134 & 33.04 & \textbf{6.23} \\
    \midrule
    FairGrad & 75.72 & 93.68 & 0.0134 & 32.25 & 5.18 \\
    Fair-FairGrad & 75.50 & 93.51 & 0.0131 & 32.54 & \textbf{4.92} \\
    \bottomrule
  \end{tabular}
  \end{adjustbox}
\end{sc}
\end{small}
\end{center}
\end{table}

\section{Theoretical Analysis}

In this section, we provide a theoretical analysis of our method on the convergence to a Pareto stationary point, at which some convex combination of task gradients is 0. As mentioned before, we assume that the gradients of different tasks are linearly independent when not reaching a Pareto stationary point. Formally, we make the following assumption, as also adopted by~\cite{navon2022multi}.

\begin{assumption}
\label{ass:gradients}
For the output sequence $\{\theta_t\}$ generated by the proposed method, the gradients of all tasks are linearly independent while not at a Pareto stationary point. 
\end{assumption}
The following assumption imposes differentiability and Lipschitz continuity on the loss functions, as also adopted by~\cite{liu2021conflict,navon2022multi}.

\begin{assumption}
\label{ass:smooth}
For each task, the loss function $l_i(\theta)$ is differentiable and $L$-smooth such that $\|\nabla l_i(\theta_1)-\nabla l_i(\theta_2)\|\leq L\|\theta_1-\theta_2\|$ for any two points $\theta_1,\theta_2$.
\end{assumption}
Then, we obtain the following convergence theorem.

\begin{theorem}
\label{thm:alpha_fair_convergence}
Suppose Assumptions \ref{ass:gradients}-\ref{ass:smooth} are satisfied. Set the stepsize $\eta_t=\frac{\sum_i w_{t,i}^{-1/\alpha}}{LK\sum_i w_{t,i}^{1-1/\alpha}}$, 
Then, there exists a subsequence $\{\theta_{t_j}\}$ of the output sequence $\{\theta_t\}$ that converges to a Pareto stationary point $\theta^\star$.
\end{theorem}
\begin{proof}[Proof skecth]
We first show that the average loss $\mathcal{L}(\theta_t)=\frac{1}{K}\sum_i l_i(\theta_t)$ is monotonically decreasing. Then, we show that the smallest singular value of the Gram matrix, denoted as $\sigma_K(G(\theta_t)^\top G(\theta_t))$, is upper bounded and approaches 0 as the number of training steps increases. Consequently, the output sequence $\{\theta_t\}$ has a subsequence converging to a point $\theta^\star$, where the matrix $G(\theta^\star)^\top G(\theta^\star)$ has a zero singular value and hence the gradients of all tasks are linearly dependent. This immediately indicates the attainment of a Pareto stationary point.
\end{proof}

\section{Conclusion}
We first discuss the connection between MTL and fair resource allocation in communication networks and model the optimization of MTL as a utility maximization problem by leveraging the concept of $\alpha$-fairness. Then, we introduce FairGrad, a novel MTL method offering the flexibility to balance different tasks through different selections of $\alpha$, and provide it with a theoretical convergence analysis. Our extensive experiments demonstrate not only the promising performance of FairGrad, but also the power of the $\alpha$-fairness idea in enhancing existing MTL methods. 

For future studies, we will explore the performance of FairGrad in more challenging MTL settings with significantly diverse tasks. Theoretically, we will study the impact of varying levels of difficulty across tasks on the final convergence and generalization performance. 




\section*{Impact Statement}

This paper discusses the fairness in optimization methods for multi-task learning (MTL).  There are some potential societal consequences, none of which we feel must be specifically highlighted here.

\nocite{langley00}

\bibliography{reference}
\bibliographystyle{icml2024}

\newpage
\appendix
\onecolumn
\section{Experiments}
\label{app:experiments}

\subsection{Toy Example}
\label{app:toy}

Following \cite{navon2022multi,liu2023famo}, we use a slightly modified version of the 2-task toy example provided in \cite{liu2021conflict}. The two tasks $L_1(x)$ and $L_2(x)$ are defined on $x=(x_1,x_2)^\top\in\mathbb{R}^2$,
\begin{align*}
    L_1(x) &= 0.1\cdot (f_1(x)g_1(x)+f_2(x)h_1(x))\\
    L_2(x) &= f_1(x)g_2(x)+f_2(x)h_2(x),
\end{align*}
where the functions are given by 
\begin{align*}
    f_1(x) &= \max\bigl(\tanh(0.5x_2),0\bigr)\\
    f_2(x) &= \max\bigl(\tanh(-0.5x_2),0\bigr)\\
    g_1(x) &= \log\Bigl(\max\bigl(|0.5(-x_1-7)-\tanh(-x_2)|,0.000005\bigr)\Bigr) + 6\\
    g_2(x) &= \log\Bigl(\max\bigl(|0.5(-x_1+3)-\tanh(-x_2)+2|,0.000005\bigr)\Bigr) + 6\\
    h_1(x) &= \bigl((-x_1+7)^2 + 0.1(-x_1-8)^2\bigr)/10 - 20\\
    h_2(x) &= \bigl((-x_1-7)^2 + 0.1(-x_1-8)^2\bigr)/10 - 20.
\end{align*}

The magnitude of the gradient of $L_2(x)$ is larger than $L_1(x)$, posing challenges to the optimization of MTL methods. By choosing different values of $\alpha$, our method covers different ideas of fairness. We use five different starting points $\{(-8.5, 7.5), (0, 0), (9.0,9.0), (-7.5,-0.5), (9.0,-1.0)\}$. We use Adam optimizer with a learning rate of 1e-3. The training process stops when the Pareto front is reached. The optimization trajectories illustrated in \cref{fig:toy_example} demonstrate that the proposed FairGrad can not only converge to the Pareto front but also exhibit different types of fairness under different choices of $\alpha$.

\subsection{Detailed Results on Multi-Task Regression}

We provide more details about per-task results on the QM9 dataset in \cref{tab:full_qm9}. Our FairGrad obtains the best $\Delta m\%$. In addition, as a special case of $\alpha$-fair loss transformation, SI outperforms other methods in 7 tasks, indicating the effectiveness of the transformation.

\begin{table}[H]
\caption{Detailed results of  on QM9 (11-task) dataset. Each experiment is repeated 3 times with different random seeds and the average is reported.}
\label{tab:full_qm9}
\begin{center}
\begin{small}
\begin{sc}
\begin{adjustbox}{max width=\textwidth}
  \begin{tabular}{llllllllllllll}
    \toprule
    \multirow{2}*{Method} & $\mu$ & $\alpha$ & $\epsilon_{HOMO}$ & $\epsilon_{LUMO}$ & $\langle R^2\rangle$ & ZPVE & $U_0$ & $U$ & $H$ & $G$ & $c_v$ & \multirow{2}*{MR$\downarrow$} & \multirow{2}*{$\Delta m\%\downarrow$} \\
    \cmidrule(lr){2-12}
    & \multicolumn{11}{c}{MAE $\downarrow$} & & \\
    \midrule
    STL & 0.067 & 0.181 & 60.57 & 53.91 & 0.502 & 4.53 & 58.8 & 64.2 & 63.8 & 66.2 & 0.072 & & \\
    \midrule
    LS & 0.106 & 0.325 & \textbf{73.57} & 89.67 & 5.19 & 14.06 & 143.4 & 144.2 & 144.6 & 140.3  & 0.128 & 8.18 & 177.6 \\
    SI & 0.309 & 0.345 & 149.8 & 135.7 & \textbf{1.00} & \textbf{4.50} & \textbf{55.3} & \textbf{55.75} & \textbf{55.82} & \textbf{55.27}  & 0.112 & 4.82 & 77.8 \\
    RLW & 0.113 & 0.340 & 76.95 & 92.76 & 5.86 & 15.46 & 156.3 & 157.1 & 157.6 & 153.0  & 0.137 & 9.55 & 203.8 \\
    DWA & 0.107 & 0.325 & 74.06 & 90.61 & 5.09 & 13.99 & 142.3 & 143.0 & 143.4 & 139.3  & 0.125 & 7.82 & 175.3 \\
    UW & 0.386 & 0.425 & 166.2 & 155.8 & 1.06 & 4.99 & 66.4 & 66.78 & 66.80 & 66.24  & 0.122 & 6.18 & 108.0 \\
    MGDA & 0.217 & 0.368 & 126.8 & 104.6 & 3.22 & 5.69 & 88.37 & 89.4 & 89.32 & 88.01  & 0.120 & 7.73 & 120.5 \\
    PCGrad & 0.106 & 0.293 & 75.85 & 88.33 & 3.94 & 9.15 & 116.36 & 116.8 & 117.2 & 114.5  & 0.110 & 6.36 & 125.7 \\
    CAGrad & 0.118 & 0.321 & 83.51 & 94.81 & 3.21 & 6.93 & 113.99 & 114.3 & 114.5 & 112.3  & 0.116 & 7.18 & 112.8 \\
    IMTL-G & 0.136 & 0.287 & 98.31 & 93.96 & 1.75 & 5.69 & 101.4 & 102.4 & 102.0 & 100.1  & 0.096 & 6.09 & 77.2 \\
    Nash-MTL & \textbf{0.102} & \textbf{0.248} & 82.95 & \textbf{81.89} & 2.42 & 5.38 & 74.5 & 75.02 & 75.10 & 74.16  & \textbf{0.093} & \textbf{3.64} & 62.0 \\
    FAMO & 0.15 & 0.30 & 94.0 & 95.2 & 1.63 & 4.95 & 70.82 & 71.2 & 71.2 & 70.3 & 0.10 & 4.73 & 58.5 \\
    \midrule
    FairGrad & 0.117 & 0.253 & 87.57 & 84.00 & 2.15 & 5.07 & 70.89 & 71.17 & 71.21 & 70.88 & 0.095 & 3.82 & \textbf{57.9} \\
    \bottomrule
  \end{tabular}
  \end{adjustbox}
\end{sc}
\end{small}
\end{center}
\vskip -0.1in
\end{table}

\subsection{Detailed Results on Effect of Different Fairness Criteria}\label{app:fairness_effect}

We provide additional results of different fairness criteria discussed in \cref{exp:fairness} in \cref{tab:supp_cityscapes} and \cref{tab:supp_nyuv2}. As $\alpha$ changes, the algorithm will prioritize certain tasks over others. Hence, an overall performance drop may be observed. However, tasks with lower priority may get higher ranks, then leading to a higher MR value.

\begin{table}[H]
\caption{Results of Different Fairness Criteria on Cityscapes (2-task) dataset. Each experiment is repeated 3 times with different random seeds and the average is reported.}
\label{tab:supp_cityscapes}
\vskip 0.15in
\begin{center}
\begin{small}
\begin{sc}
\begin{adjustbox}{max width=\textwidth}
  \begin{tabular}{lllllll}
    \toprule
    \multirow{2}*{Method} & \multicolumn{2}{c}{Segmentation} & \multicolumn{2}{c}{Depth} & 
    \multirow{2}*{MR $\downarrow$} &
    \multirow{2}*{$\Delta m\%\downarrow$} \\
    \cmidrule(lr){2-3}\cmidrule(lr){4-5}
    & mIoU $\uparrow$ & Pix Acc $\uparrow$ & Abs Err $\downarrow$ & Rel Err $\downarrow$ & \\
    \midrule
    FairGrad ($\alpha=1$) & \textbf{75.94} & \textbf{93.65} & 0.0138 & 33.13 & \textbf{2.25} & 6.73 \\
    FairGrad ($\alpha=2$) & 74.10 & 93.03 & 0.0135 & 29.92 & 5.75 & \textbf{3.90} \\
    FairGrad ($\alpha=5$) & 67.30 & 90.23 & \textbf{0.0134} & 30.01 & 7.25 & 6.87 \\
    FairGrad ($\alpha=10$) & 62.77 & 88.00 & 0.0151 & \textbf{28.05} & 8.50 & 10.54 \\
    \bottomrule
  \end{tabular}
  \end{adjustbox}
\end{sc}
\end{small}
\end{center}
\vskip -0.1in
\end{table}

\begin{table}[H]
\caption{Results of Different Fairness Criteria on NYU-v2 (3-task) dataset. Each experiment is repeated 3 times with different random seeds and the average is reported.}
\label{tab:supp_nyuv2}
\vskip 0.15in
\begin{center}
\begin{small}
\begin{sc}
\begin{adjustbox}{max width=\textwidth}
  \begin{tabular}{llllllllllll}
    \toprule
    \multirow{3}*{Method} & \multicolumn{2}{c}{Segmentation} & \multicolumn{2}{c}{Depth} & \multicolumn{5}{c}{Surface Normal} & \multirow{3}*{MR $\downarrow$} & \multirow{3}*{$\Delta m\%\downarrow$} \\
    \cmidrule(lr){2-3}\cmidrule(lr){4-5}\cmidrule(lr){6-10}
    & \multirow{2}*{mIoU $\uparrow$} & \multirow{2}*{Pix Acc $\uparrow$} & \multirow{2}*{Abs Err $\downarrow$} & \multirow{2}*{Rel Err $\downarrow$} & \multicolumn{2}{c}{Angle Distance $\downarrow$} & \multicolumn{3}{c}{Within $t^\circ$ $\uparrow$} & \\
    \cmidrule(lr){6-7}\cmidrule(lr){8-10}
    & & & & & Mean & Median & 11.25 & 22.5 & 30 & \\
    \midrule
    FairGrad ($\alpha=1$) & \textbf{40.64} & \textbf{67.20} & 0.5671 & 0.2434 & 25.18 & 20.05 & 28.35 & 55.61 & 68.37 & \textbf{4.78} & -2.79 \\
    FairGrad ($\alpha=2$) & 38.80 & 65.29 & \textbf{0.5572} & \textbf{0.2322} & 24.55 & 18.97 & 30.50 & 57.94 & 70.14 & \textbf{4.78} & \textbf{-4.96} \\
    FairGrad ($\alpha=5$) & 34.05 & 62.82 & 0.5853 & 0.2375 & \textbf{24.40} & \textbf{18.70} & \textbf{30.96} & \textbf{58.48} & \textbf{70.45} & 6.22 & -3.03 \\
    FairGrad ($\alpha=10$) & 31.45 & 61.43 & 0.5948 & 0.2341 & 24.80 & 19.08 & 30.10 & 57.58 & 69.69 & 6.22 & -1.00 \\
    \bottomrule
  \end{tabular}
  \end{adjustbox}
\end{sc}
\end{small}
\end{center}
\vskip -0.1in
\end{table}

\begin{table}[H]
\caption{Results of Different Methods for Solving Sub-problems of Nash-MTL on Cityscapes (2-task) dataset. Each experiment is repeated 3 times with different random seeds and the average is reported.}
\label{tab:supp_nashmtl}
\vskip 0.15in
\begin{center}
\begin{small}
\begin{sc}
\begin{adjustbox}{max width=\textwidth}
  \begin{tabular}{llllll}
    \toprule
    \multirow{2}*{Method} & \multicolumn{2}{c}{Segmentation} & \multicolumn{2}{c}{Depth} & 
    \multirow{2}*{$\Delta m\%\downarrow$}  \\
    \cmidrule(lr){2-3}\cmidrule(lr){4-5}
    & mIoU $\uparrow$ & Pix Acc $\uparrow$ & Abs Err $\downarrow$ & Rel Err $\downarrow$ & \\
    \midrule
    Nash-MTL (original) & 75.41 & 93.66 & 0.0129 & 35.02 & 6.82 \\
    Nash-MTL (ours) & 75.26 & 93.71 & 0.0129 & 34.45 & \textbf{6.28} \\
    \bottomrule
  \end{tabular}
  \end{adjustbox}
\end{sc}
\end{small}
\end{center}
\vskip -0.1in
\end{table}

\subsection{Difference Between FairGrad and Nash-MTL}
Although FairGrad with proportional fairness shares similarities with Nash-MTL, there are many differences. The high-level ideas are different. Nash-MTL is developed from the perspective of game theory, whereas our FairGrad is inspired by fair resource allocation in communication networks. Thus, this allows us to incorporate the advances from network resource allocation into MTL.
FairGrad incorporates other different notions of fairness that Nash-MTL cannot cover. From our empirical studies on Cityscapes and NYUv2 datasets, the performance of MDP fairness is significantly better than proportional fairness. This indicates that proportional fairness may not always be the most suitable choice for different applications and scenarios. This greatly highlights the importance of incorporating other fairness ideas. Unlike Nash-MTL, our proposed FairGrad offers the flexibility to explore different fairness criteria.

Technically, algorithmic designs are different. We propose to solve weights ${w_1,...,w_K}$ from our objective through a simple nonlinear least square problem. Solving this problem is efficient, and it turns out that the results are good enough. As a comparison, Nash-MTL solves weights from a different constrained objective via a variation of concave-convex-procedure (CCP) that solves a sequence of simple constrained problems. Our approach that solves a nonlinear least square equation can also be applied to Nash-MTL. 

We experiment on the Cityscapes dataset using the codes from the Nash-MTL paper. The results are presented in \cref{tab:supp_nashmtl}, where Nash-MTL (original) denotes the original method, and Nash-MTL (ours) denotes the method using our approach to solve the sub-problems. The results demonstrated that our approach can not only be applied to Nash-MTL, but also achieve slightly better performance.


\section{Proofs}
\label{app:proofs}

\subsection{$\alpha$-fairness}\label{diff_alpha}
Different values of $\alpha$ yield different ideas of fairness. Recall from \cref{eq:alpha_fair_utility} the following utilization maximization objective
\begin{align*}
  \max_{x_1,...,x_K\in\mathcal{D}} \quad \sum_{i\in[K]}u(x_i):=\frac{x_i^{1-\alpha}}{1-\alpha},
  \end{align*}
where $x_i$ denotes the transmission rate of user $i$, $u(x_i)=\frac{x_i^{1-\alpha}}{1-\alpha}$ is a concave utility function with $\alpha\in[0,1)\cup(1,+\infty)$, and $\mathcal{D}$ is the convex link capacity constraints.

When $\alpha=0$, the objective is 
\begin{align*}
\begin{split}
     \max_{x_1,...,x_K\in\mathcal{D}} \quad \sum_{i\in[K]}u(x_i):=x_i,
\end{split}
\end{align*}
Note that in our MTL setting, it is similar to the Linear Scalarization.   

When $\alpha\to 1$, the utilization maximization objective \cref{eq:alpha_fair_utility} captures the proportional fairness. First note that
\begin{align*}
   \max_{x_1,...,x_K\in\mathcal{D}}\quad \sum_{i\in[K]} \frac{x_i^{1-\alpha}}{1-\alpha}= \max_{x_1,...,x_K\in\mathcal{D}} \quad \sum_{i\in[K]} \frac{x_i^{1-\alpha}-1}{1-\alpha}.
\end{align*}
By applying L'Hospital's rule, we have
\begin{align*}
    \lim_{\alpha\to 1} \frac{x_i^{1-\alpha}-1}{1-\alpha}=\log x_i.
\end{align*}
Then, the current objective is
\begin{align*}
\begin{split}
   \max_{x_1,...,x_K\in\mathcal{D}} \quad \sum_{i\in[K]}\log x_i.
\end{split}
\end{align*}
For a concave function $f(x)$ over a domain $\mathcal{D}$, it is shown in \cite{srikant2013communication} that
\begin{align}
    \nabla f(x^\star)(x-x^\star)\le0\quad \forall x\in\mathcal{D}.
\label{ineq:concave}
\end{align}
Clearly, since the objective $\sum_{i\in[K]} \log x_i$ is concave, applying \cref{ineq:concave} yields
\begin{align*}
    \sum_{i\in[K]} \frac{x_i-x_i^\star}{x_i^\star}\le0.
\end{align*}
If the proportion of one user increases, then there will be at least one other user whose proportional change decreases. The allocation $\{x^\star\}$ captures the proportional fairness.
In the MTL setting, the objective becomes
\begin{align*}
\begin{split}
    &\max_{d\in B_\epsilon}\quad  \sum_{i\in[K]} \log g_i^\top d\\
    &\;\;\text{s.t.}\quad g_i^\top d\ge0,
\end{split}
\end{align*}
which corresponds to Nash-MTL~\cite{navon2022multi}.

When $\alpha\to \infty$, the utilization maximization objective \cref{eq:alpha_fair_utility} yields the max-min fairness. The following proofs follow Section 2.2.1 in \cite{srikant2013communication}. Let $x^\star(\alpha)$ be the $\alpha$-fair allocation. Assume $x_i^\star(\alpha)\to x^\star$ as $\alpha\to \infty$ and $x_1^\star<x_2^\star<\cdots <x_K^\star$. Let $\epsilon$ be the minimum difference of $\{x^\star\}$. That is, $\epsilon=\min_i |x_{i+1}^\star-x_i^\star|$, $i\in[K-1]$. When $\alpha$ is sufficiently large, we then have $|x_i^\star(\alpha)-x_i^\star|\le\epsilon/4$, which also implies $x_1^\star(\alpha)<x_2^\star(\alpha)<\cdots <x_K^\star(\alpha)$. According to \cref{ineq:concave}, we have
\begin{align*}
    \sum_{i\in[K]} \frac{x_i-x_i^\star(\alpha)}{x_i^{\star\alpha}(\alpha)}\le0.
\end{align*}
For any $j\in[K]$, the following inequality always holds
\begin{align*}
    \sum_{i=1}^j (x_i-x_i^\star(\alpha))\frac{x_j^{\star\alpha}(\alpha)}{x_i^{\star\alpha}(\alpha)} + (x_j-x_j^\star(\alpha)) + \sum_{i=j+1}^K (x_i-x_i^\star(\alpha))\frac{x_j^{\star\alpha}(\alpha)}{x_i^{\star\alpha}(\alpha)}\le0.
\end{align*}
Since we have $|x_i^\star(\alpha)-x_i^\star|\le\epsilon/4$, we then get
\begin{align*}
    \sum_{i=1}^j (x_i-x_i^\star(\alpha))\frac{x_j^{\star\alpha}(\alpha)}{x_i^{\star\alpha}(\alpha)} + (x_j-x_j^\star(\alpha)) - \sum_{i=j+1}^K |x_i-x_i^\star(\alpha)|\frac{(x_j^\star+\epsilon/4)^\alpha}{(x_i^\star-\epsilon/4)^\alpha}\le0,
\end{align*}
where $(x_i-\epsilon/4)-(x_j^\star+\epsilon/4)\ge\epsilon/2$ for any $i>j$. Therefore, when $\alpha$ becomes large enough, the last term in the above inequality will be negligible. Consequently, if $x_j>x_j^\star(\alpha)$, then the allocation for at least one user $i<j$ will decrease. That is, the allocation approaches the max-min fairness when $\alpha\to \infty$. In the context of MTL, this takes the same spirit as MGDA and its variants that aim to maximize the loss decrease for the least-fortune task. 

\subsection{Convergence}
\begin{theorem}[Restatement of \cref{thm:alpha_fair_convergence}]
Suppose Assumptions \ref{ass:gradients}-\ref{ass:smooth} are satisfied. Set the stepsize $\eta_t=\frac{\sum_i w_{t,i}^{-1/\alpha}}{LK\sum_i w_{t,i}^{1-1/\alpha}}$, 
Then, there exists a subsequence $\{\theta_{t_j}\}$ of the output sequence $\{\theta_t\}$ that converges to a Pareto stationary point $\theta^\star$.
\end{theorem}

\begin{proof}
Since $(g_i^\top d)^{-\alpha}=w_i$ and $d=\sum_i w_ig_i$ in each iteration, we have the norm $\|d\|^2=\sum_i w_ig_i^\top d=\sum_i w_i^{1-\frac{1}{\alpha}}$.

Each loss function $l_i(\theta)$ is $L$-smooth. Then, we have
\begin{align*}
\begin{split}
    l_i(\theta_{t+1})&\le l_i(\theta_t) -\eta_t g_{t,i}^\top d_t + \frac{L}{2}\|\eta_t d_t\|^2\\
    &=l_i(\theta_t)-\eta_t w_{t,i}^{-\frac{1}{\alpha}} + \frac{L}{2}\eta_t^2\|d_t\|^2\\
    &=l_i(\theta_t) - \eta_t w_{t,i}^{-\frac{1}{\alpha}} + \frac{L\eta_t^2}{2}(\sum_{j=1}^K w_{t,j}^{1-\frac{1}{\alpha}}).
\end{split}
\end{align*}
Set the learning rate $\eta_t=\frac{\sum_{i=1}^Kw_{t,i}^{-1/\alpha}}{LK\sum_{i=1}^K w_{t,i}^{1-1/\alpha}}$. Consider the averaged loss function $\mathcal{L}(\theta)=\frac{1}{K}\sum_i l_i(\theta)$, we have
\begin{align*}
\begin{split}
    \mathcal{L}(\theta_{t+1})&\le\mathcal{L}(\theta_t) - \eta_t\frac{1}{K}\sum_{i=1}^K{w_{t,i}^{-\frac{1}{\alpha}}} + \frac{L\eta_t^2}{2}(\sum_{i=1}^K w_{t,i}^{1-\frac{1}{\alpha}})\\
    &=\mathcal{L}(\theta_t) - L\eta_t^2(\sum_{i=1}^K w_{t,i}^{1-\frac{1}{\alpha}}) + \frac{L\eta_t^2}{2}(\sum_{i=1}^K w_{t,i}^{1-\frac{1}{\alpha}})\\
    &=\mathcal{L}(\theta_t) - \frac{L\eta_t^2}{2}(\sum_{i=1}^K w_{t,i}^{1-\frac{1}{\alpha}}).
\end{split}
\end{align*}
It can be observed that $\sum_{\tau=0}^t \frac{L\eta_\tau^2}{2}(\sum_{i=1}^K w_{\tau,i}^{1-\frac{1}{\alpha}}) \le\mathcal{L}(\theta_0)-\mathcal{L}(\theta_{t+1})$. Then, we get
\begin{align*}
    \sum_{\tau=0}^\infty \frac{L\eta_\tau^2}{2}(\sum_{i=1}^K w_{\tau,i}^{1-\frac{1}{\alpha}}) = \frac{1}{2LK^2}\sum_{\tau=0}^\infty \frac{(\sum_{i=1}^K w_{\tau,i}^{-\frac{1}{\alpha}})^2}{\sum_{i=1}^K w_{\tau,i}^{1-\frac{1}{\alpha}}}<\infty.
\end{align*}
Then, it can be obtained that
\begin{align}
    \lim_{\tau\to\infty} \frac{(\sum_{i=1}^K w_{\tau,i}^{-\frac{1}{\alpha}})^2}{\sum_{i=1}^K w_{\tau,i}^{1-\frac{1}{\alpha}}}=0.
\label{eq:limit}
\end{align}

From \cref{eq:alpha_fair_weight}, we get
\begin{align*}
    \|w_{t}^{-\frac{1}{\alpha}}\|\ge\sigma_K(G_t^\top G_t)\|w_t\|,
\end{align*}
where $\sigma_K(G_t^\top G_t)$ is the smallest singular value of matrix $G_t^\top G_t$. Denote $\mathbf{1}=[1,\cdots,1]^\top$ as the length-$K$ vector whose elements are all 1. Note that we have
\begin{align*}
    \|w\|^2=\sum_{i=1}^K w_i^2\le\sum_{i=1}^K w_i\cdot\sum_{i=1}^K w_i=\|w\|_1^2,
\end{align*}

\begin{align*}
    \|w\|_1=\mathbf{1}^\top w\le\|1\|\cdot\|w\|=\sqrt{K}\|w\|.
\end{align*}

Combine the above inequalities, we get
\begin{align*}
    \|w_t^{-\frac{1}{\alpha}}\|_1\ge\|w_{t}^{-\frac{1}{\alpha}}\|\ge\sigma_K(G_t^\top G_t)\|w_t\|\ge\frac{1}{\sqrt{K}}\sigma_K(G_t^\top G_t)\|w_t\|_1.
\end{align*}
Then, we have
\begin{align}
    \frac{\sum_{i=1}^K w_{t,i}^{-\frac{1}{\alpha}}}{\sum_{i=1}^K w_{t,i}}\ge\frac{1}{\sqrt{K}}\sigma_K(G_t^\top G_t).
\label{ineq:singular_value}
\end{align}
Furthermore,
\begin{align}
\begin{split}
    \frac{\sum_{i=1}^K w_{t,i}^{-\frac{1}{\alpha}}}{\sum_{i=1}^K w_{t,i}}&=\frac{(\sum_{i=1}^K w_{t,i}^{-\frac{1}{\alpha}})^2}{(\sum_{i=1}^K w_{t,i})\cdot(\sum_{i=1}^K w_{t,i}^{-\frac{1}{\alpha}})}\\
    &=\frac{(\sum_{i=1}^K w_{t,i}^{-\frac{1}{\alpha}})^2}{\sum_{i=1}^K w_{t,i}^{1-\frac{1}{\alpha}} + \sum_{i=1}^K\sum_{j=1,j\neq i}^K w_{t,i}w_{t,j}^{-\frac{1}{\alpha}}}\\
    &\le\frac{(\sum_{i=1}^K w_{t,i}^{-\frac{1}{\alpha}})^2}{\sum_{i=1}^K w_{t,i}^{1-\frac{1}{\alpha}}}.
\end{split}
\label{ineq:singular_value2}
\end{align}

For any fixed $K$, it can be concluded from \cref{eq:limit}, \cref{ineq:singular_value}, and \cref{ineq:singular_value2} that
\begin{align*}
    \lim_{\tau\to\infty} \sigma_K(G_\tau^\top G_\tau)=0.
\end{align*}
Since the sequence $\mathcal{L}(\theta_t)$ is monotonically decreasing, we know the sequence $\theta_t$ is in the compact sublevel set $\{\theta|\mathcal{L}(\theta)\le\mathcal{L}(\theta_0)\}$. Then, there exists a subsequence $\theta_{t_j}$ that converges to $\theta^\star$ where we have $\sigma_K(G_\star^\top G_\star)=0$ and $G_\star$ denotes the matrix of multiple gradients at $\theta^\star$. Therefore, the gradients at $\theta^\star$ are linearly dependent, and $\theta^\star$ is Pareto stationary.
\end{proof}

\subsection{$\alpha$-fair loss transformation}
\label{alpha_fair_loss}
\begin{proposition}[Restatement of \cref{prop:alpha_fair_loss}]
    The Pareto front of the $\alpha$-fair loss functions in \cref{alpha_trans} is the same as that of original loss functions $(l_1,...,l_K)$.
\end{proposition}

\begin{proof}
If $\theta^\star$ is a Pareto optimal point of $L(\theta)$, then there exists no point $\theta$ dominating $\theta^\star$. That is, we have $l_i(\theta^\star)\le l_i(\theta)$ for all $i\in[K]$ and $L(\theta^\star)\neq L(\theta)$. Note that the function $f(x)=\frac{x^{1-\alpha}}{1-\alpha}$ with $x>0$ and $\alpha\in[0,1)\cup(1,+\infty)$ is monotonically increasing. It is evident that $\frac{l_i^{1-\alpha}(\theta^\star)}{1-\alpha}\le\frac{l_i^{1-\alpha}(\theta)}{1-\alpha}$ for all $i\in[K]$. Thus, $\theta^\star$ is also a Pareto optimal point of $\frac{L^{1-\alpha}(\theta)}{1-\alpha}$. Similarly, it can be shown that if $\theta^\star$ is a Pareto optimal point of $\frac{L^{1-\alpha}(\theta)}{1-\alpha}$, it is also a Pareto optimal point of $L(\theta)$.
\end{proof}

\end{document}